\documentclass[journal, onecolumn]{IEEEtran}

\usepackage{src/dependencies}

\title{Deep ARTMAP: Generalized Hierarchical Learning with Adaptive Resonance Theory}
\author{
    \IEEEauthorblockN{
    Niklas~M.~Melton~\orcidlink{0000-0001-9625-7086}\IEEEauthorrefmark{1}, 
        Leonardo~Enzo~Brito~da~Silva~\orcidlink{0000-0003-3639-9175}\IEEEauthorrefmark{2}, 
        Sasha~Petrenko~\orcidlink{0000-0003-2442-8901}\IEEEauthorrefmark{1}\IEEEauthorrefmark{2},
        Donald~C.~Wunsch~II~\orcidlink{0000-0002-9726-9051}\IEEEauthorrefmark{1}\IEEEauthorrefmark{2}} \\
    \IEEEauthorblockA{
        \IEEEauthorrefmark{1}Applied Computational Intelligence Lab, \\
        Missouri University of Science and Technology, Rolla, Missouri, USA \\
        niklasmelton@mst.edu, petrenkos@mst.edu, dwunsch@mst.edu
    } \\
    \IEEEauthorblockA{
        \IEEEauthorrefmark{2}Kummer Institute Center for Artificial Intelligence and Autonomous Systems, \\
        Missouri University of Science and Technology, Rolla, Missouri, USA \\
        lbritodasilva@mst.edu
    }
}
\date{November 2024}

\begin{document}

\maketitle

\begin{abstract}
This paper presents \method, a novel extension of the ARTMAP architecture that generalizes the self-consistent modular ART (SMART) architecture to enable hierarchical learning (supervised and unsupervised) across arbitrary transformations of data.
The \method framework operates as a divisive clustering mechanism, supporting an arbitrary number of modules with customizable granularity within each module. Inter-ART modules regulate the clustering at each layer, permitting unsupervised learning while enforcing a one-to-many mapping from clusters in one layer to the next.
While \method reduces to both ARTMAP and SMART in particular configurations, it offers significantly enhanced flexibility, accommodating a broader range of data transformations and learning modalities.

\end{abstract}

\section{Introduction}
Numerous ART and ARTMAP applications incorporate hierarchical clustering methods \cite{Bartfai1994HierarchicalCW, wunsch1993optoelectronic, ishihara1995arboart, bartfai1996art, bartfai1997adaptive, bartfai1997fuzzy, da2020distributed} and multi-channel data fusion \cite{tan2007intelligence, meng2013semi, meng2012heterogeneous}.
However, current architectures lack the capacity to uncover hierarchical structures within multi-modal data.
Consider the problem of finding relationships between common animal names, animal photos, and animal calls.
For example, the word ``bird" can correspond to numerous images of various bird species and each image can correspond to numerous audio samples of each species' distinct songs: $audio~samples \subseteq bird~images \subseteq word~bird$.
The songs and images all belong to the class ``bird" but there is a one-to-many mapping from ``bird" to images and from each image to a subset of songs.
Such data presents a multi-modal format with an embedded hierarchical relationship.
\method is designed to bridge this gap, providing a powerful tool for multi-modal hierarchical data clustering applications.
\method inherits key properties from both ARTMAP and SMART, including adaptive learning, resistance to catastrophic forgetting, and hierarchical self-consistency.
\method reduces to SMART with the use of identity transforms and ARTMAP with the use of a single module, and it extends their capabilities by enabling supervised learning within SMART, supporting multi-layer configurations within ARTMAP, such that $\text{ARTMAP} \subseteq \text{SMART} \subseteq \text{\method}$.

\section{Background}
\subsection{Adaptive Resonance Theory (ART)}
Adaptive Resonance Theory (ART) is a theory of learning in mammalian brains that has inspired a family of unsupervised clustering algorithms that address the stability-plasticity dilemma \cite{Carpenter1988AMP, Grossberg2013AdaptiveRT, BRITODASILVA2019167}.
These algorithms are adaptive in the number of categories they create and are typically immune to catastrophic forgetting.
The core principals common to all ART algorithms are the activation and match-criterion functions.
The activation function scores candidate clusters against each sample presentation.
Cluster candidates are then sorted according to their activation and each candidate is evaluated one at a time using the match-criterion function.
The match-criterion function regulates the granularity of the category prototypes, thereby balancing under- and over- fitting of the data.
A threshold, known as the vigilance parameter $\rho$, is checked against the match-criterion value for the candidate and determines if the candidate cluster is in the resonant state.
If the candidate cluster is resonant (i.e. passes the vigilance check) the cluster is updated to incorporate the sample.
If, on the other hand, the cluster is not resonant, the activation of the cluster is suppressed and the search for a resonant cluster continues.
This process repeats until either a resonant cluster is found or no clusters remain activated.
When no clusters remain, a new cluster is created from the sample.
This process allows the vigilance parameter to control the granularity of the clustering without requiring the user to specify the number of clusters explicitly.

\subsection{ARTMAP}
\label{ss:artmap}

The ARTMAP~\cite{Carpenter1991ARTMAPSR} algorithm permits supervised learning using unsupervised ART networks~\cite{Carpenter1988AMP, Grossberg2013AdaptiveRT, BRITODASILVA2019167} as the foundation.
ARTMAP uses a pair of ART modules (A and B) to cluster the dependent and independent data streams separately with an inter-ART map field enforcing a one-to-many mapping from B- to A-side clusters.
The B-side receives samples from the dependent data and proceeds to cluster in an unsupervised fashion.
The A-side, however, receives samples from the independent data and also clusters in an unsupervised fashion unless the A-side cluster maps (via the inter-ART module) to a B-side cluster which does not coincide with the corresponding dependent data sample. When this occurs, a match-tracking procedure\footnote{See~\cite{matchTracking2025} and the references therein.} activates which raises the vigilance of the A-side module such that the A-side cluster can no longer resonate and the search for a resonant A-side cluster continues.
In doing so, ARTMAP compresses the representation of both data streams and learns the functional mapping from A-clusters to B-clusters.

\subsection{Simplified ARTMAP}
\label{ss:simplified-artmap}

Simplified ARTMAP~\cite{kasuba1993simplified,gotarredona1998adaptive} was introduced to simplify ARTMAP learning for the classification of integer-valued labels.
While ARTMAP is capable of classification, the overhead of clustering class labels is not strictly necessary when they exist as discrete labels such as integers.
For such situations, it is possible to replace the B-side of ARTMAP with the class labels.
This eliminates the need for a B-side ART module and permits computational resources to be concentrated on the A-side module.

\subsection{Self-consistent Modular ART (SMART)}
SMART~\cite{Bartfai1994HierarchicalCW} is an extension of ARTMAP that generates an unsupervised hierarchical clustering of a single data stream.
SMART allows an arbitrary number of modules rather than the original A and B sides only, thereby learning the mapping from each module to the next.
A monotonically increasing vigilance value for each subsequent layer ensures that SMART generates a divisive hierarchy of categories; with the least number of categories corresponding to the layer with the lowest vigilance (root module) and the largest number of categories belonging to the layer with the highest vigilance (leaf module).
In other words, in SMART, a module with a smaller vigilance parameter value drives the categorization of the subsequent module with a larger vigilance parameter value.
The hierarchy remains self-consistent as each category in module $\LAYERINDEX$ is split into 1 or more categories in module $\LAYERINDEX + 1$ with no overlap in child categories between the parent categories.

\section{D\MakeLowercase{eep}ARTMAP}
\method (Fig.~\ref{fig:deepartmap-tikz}) is a generalization of SMART that extends its functionality significantly by augmenting it with:

\begin{enumerate}
    \item \textbf{Supervised Learning Ability}: \method extends the unsupervised clustering mechanisms of SMART to optionally support supervised learning.
    Both SMART and the unsupervised variant of \method can be understood as a single ARTMAP driving a chain of $\NMODULES-2$ Simplified ARTMAPs below it, where $\NMODULES$ is the number of hierarchical modules.
    In contrast, the supervised variant of \method can be understood as a chain of $\NMODULES$ simplified ARTMAP with the B-side of the top-most simplified ARTMAPs being driven directly by the class label vector.
    While Simplified ARTMAP does not require the Inter-ART map field, the Simplified ARTMAPs in our implementation do use Inter-ART map fields similar to standard ARTMAP in order to permit full functionality. 
    These simple changes permits the user the option of finding hierarchical clusters either with or without class labels enforcing a partition.
    \item \textbf{Arbitrary Transformations as Inputs}: While SMART uses the same input $\SAMPLE$ for all modules, \method generalizes this behavior, such that the input to its module $\LAYERINDEX$ is given by:
    \begin{equation}
        \SAMPLE_{\LAYERINDEX+1} \in \{ f_{\LAYERINDEX+1}(\SAMPLE_k) \mid f_{\LAYERINDEX+1} \text{ is any function} \}
        \label{Eq:deep_artmap_input}
    \end{equation}
    \noindent where $\SAMPLE_1$ is the original feature matrix and $f_{\LAYERINDEX+1}(\SAMPLE_k)$ is any transformation of the previous feature space. The transformation is not required to have a closed-form solution or be analytically defined. It may instead represent any arbitrary many-to-one mapping, whether concrete, implicit, or purely theoretical. This powerful generalization permits \method to self-consistently cluster on iterative transformations or abstractions of the feature space. 
\end{enumerate}

In particular, \method reduces to SMART when used in unsupervised mode and when $f_{\LAYERINDEX}(\SAMPLE)$ is the identity function for all modules $\LAYERINDEX \in \{1, \ldots, \NMODULES\}$.
Moreover, \method reduces to ARTMAP when used in supervised mode with two layers and when $f_{2}(\SAMPLE)$ is the identity function and $f_{1}$ is the target set $\LABELS$.

\subsection{Algorithm}
\label{ss:algorithm}

Figure \ref{fig:deepartmap-tikz} demonstrates the relationship between ART submodules, ARTMAP map fields, and the nonlinear transform inputs at each submodule in \method, while Algorithm \ref{a:deepartmap-training} demonstrates the incremental training and inference procedures for \method.

\method is trained through a incremental and iterative process connecting the prescribed categories of higher ART modules to the training of ARTMAP modules assembled from lower layers.
Each ART module receives a unique transformation of the same single sample, and training proceeds from ART module $\NMODULES$ down to ART module $1$.
\method therefore reduces to SMART with the identity transform $f_{\LAYERINDEX}(\SAMPLE) = \SAMPLE$, and it reduces to ARTMAP with the use of two ART modules ($L=2$).
The final module ART module $\NMODULES$ is trained with its normal ART clustering procedure to generate a prescribed category from its transformed input $f_{\NMODULES}(\SAMPLE)$.
ART module $\NMODULES-1$ is then trained use a Simplified ARTMAP procedure \cite{kasuba1993simplified} using the assembled ART module $\NMODULES-1$ and the local supervised category prescribed by $\text{ART}_{\NMODULES}(f_{\NMODULES}(\SAMPLE))$.
This procedure continues in reverse through each ART module $\LAYERINDEX$ until all modules are trained on their respective transformations of the current sample, and the procedure restarts for the subsequent sample.

The inference procedure then processes the input samples in the same manner as the training procedure, computing the feature transforms for each ART module of the input sample and then aggregating the category labels prescribed by each ART module.
The inference procedure therefore is not recursive and can be done in parallel for each module in a single forward pass.

\tikzstyle{box} = [
    rectangle,
    text centered,
    minimum height=3em,
    minimum width=3em,
    draw=black,
]

\tikzstyle{square} = [
    regular polygon,
    regular polygon sides=4,
    draw=black,
]

\tikzstyle{arrow} = [thick,->,>=stealth]

\begin{figure}
    \centering

    \begin{tikzpicture}[
        node distance=1cm,
    ]

    \node (x1) [] {$\SAMPLE_1 = f_1 \left( \SAMPLE \right)$};
    \node (art1) [box, above = of x1] {ART Module 1};
    \draw[arrow] (x1) -- (art1);
    \coordinate[above = of art1] (art1above);
    \node (map1) [box, right = 2.1cm of art1above] {Map Field 1};
    \draw[arrow] (art1) -- (art1above) -- (map1);

    \node (ldots1) [right = of map1] {$\ldots$};
    \draw[arrow] (ldots1) -- (map1);

    \coordinate[right = of ldots1] (artkabove);
    \node (artk) [box, below = of artkabove] {ART Module $\LAYERINDEX$};
    \draw[arrow] (artk) -- (artkabove) -- (ldots1);

    \node(xk) [below = of artk] {$\SAMPLE_\LAYERINDEX = f_\LAYERINDEX \left( \SAMPLE \right)$};
    \draw[arrow] (xk) -- (artk);

    \node (ldots2) [right = of artkabove] {$\ldots$};
    \draw[arrow] (artk) -- (artkabove) -- (ldots2);

    \node (map2) [box, right = of ldots2] {Map Field $\NMODULES-1$};
    \draw[arrow] (ldots2) -- (map2);

    \coordinate[right = 2.1cm of map2] (artlabove);
    \node (artl) [box, below = of artlabove] {ART Module $\NMODULES$};
    \draw[arrow] (artl) -- (artlabove) -- (map2);

    \node (xl) [below = of artl] {$\SAMPLE_{\NMODULES} = f_\NMODULES \left( \SAMPLE \right)$};
    \draw[arrow] (xl) -- (artl);
    
    \end{tikzpicture}
    
    \caption{
        The \method model.
        A single input sample is processed by differing nonlinear functions $f_\LAYERINDEX$ and presented to a set of $L$ ART modules that interact through inter-ARTMAP map fields $1, \ldots, \NMODULES - 1$.
        \method reduces to SMART in the case of identity transforms $f_\LAYERINDEX(\SAMPLE) = \SAMPLE$, and it further reduces to ARTMAP with the use of two ART modules $\NMODULES = 2$ arranged into a single ARTMAP module during.
    }
    \label{fig:deepartmap-tikz}
\end{figure}
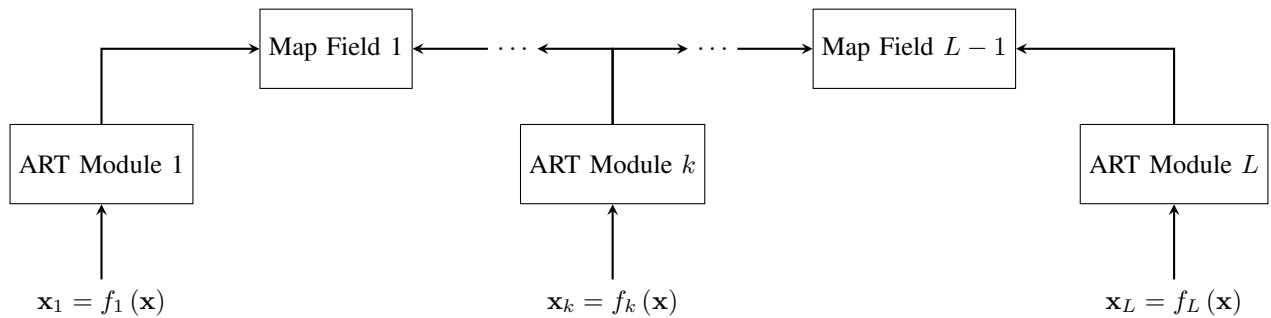

\begin{algorithm}
    \DontPrintSemicolon
    \caption{
        Online \method training and inference algorithm using an iterative Simplified ARTMAP training procedure with optional supervised labels.
        In both unsupervised and supervised scenarios, \method is trained incrementally on one sample at a time, and each ART module receives its own nonlinear transformation $f_{\LAYERINDEX}$ if the current sample $\SAMPLE$.
        The last ART module $L$ is trained using its normal ART clustering procedure to generate its category assignment; subsequent modules are trained backwards through \method using a Simplified ARTMAP procedure \cite{kasuba1993simplified} with the prescribed category of ART module $\LAYERINDEX + 1$ to train the Simplified ARTMAP module assembled from ART module and map field $\LAYERINDEX$.
        In the case that vector-valued labels are available in a supervised learning scenario, these labels are substituted for the sample input during training of ART submodule $\NMODULES$, and the training procedure above continues normally.
        In the case that integer-valued labels are available in a supervised learning scenario, the bootstrap-training of ART module $\NMODULES$ is unnecessary, and the integer label is use as the prescribed category output of ART submodule $\NMODULES$, and training continues normally thereafter.
        The inference procedure aggregates the categories prescribed by each ART module.
        When $f_{k}$ is not a callable function but a known value, the calls to the function in the algorithm can be replaced with a look-up procedure. 
    }
    \label{a:deepartmap-training}
    \KwData{
        Data samples $\SAMPLES \in \mathbb{R}^{\NSAMPLES \times \INPUTDIM}$, optional target set $\LABELS \in \mathbb{R}^{\NSAMPLES \times \LABELDIM} = f_\NMODULES (\SAMPLES)$.
    }

    \KwResult{
        \method prescribed submodule categories $\OUTPUTS=[y_{i,j}]^{\NSAMPLES \times \NMODULES}$.
    }
   
    \Comment{Notation}
    \nonl $\NSAMPLES$: number of samples.\;
    \nonl $\INPUTDIM$: sample feature dimension.\;
    \nonl $\LABELDIM$: target dimension.\;
    \nonl $\LAYERINDEX$: \method's ART module index.\;
    \nonl $\NMODULES$: number of \method ART modules.\;
    \nonl $\theta_\LAYERINDEX^{\text{ART}}$: set of hyper-parameters for the $\LAYERINDEX^{th}$ ART module within \method.\;
    \nonl $\theta_\LAYERINDEX^{\text{MF}}$: set of hyper-parameters for the $\LAYERINDEX^{th}$ Map Field module within \method.\;
    \nonl $\Theta_\LAYERINDEX=\{\theta_\LAYERINDEX^{\text{ART}}, \theta_\LAYERINDEX^{\text{MF}}\}$: set of hyper-parameters for the $\LAYERINDEX^{th}$ Simplified ARTMAP within \method.\;
    \nonl $f_\LAYERINDEX \left( \SAMPLE \right)$: function transform of sample $\SAMPLE$.\;
    \nonl $\text{ART}_\LAYERINDEX$: ART module at index $\LAYERINDEX$.\;
    \nonl $\text{MF}_\LAYERINDEX$: ARTMAP map field at index $\LAYERINDEX$.\;
    \nonl $\OUTPUT_\LAYERINDEX$: resonant category of $\text{ART}_\LAYERINDEX$. \;
    \Comment{Training}
    \For{
        $i = \{ 1, \ldots, \NSAMPLES \}$
    }{
        $\SAMPLE \gets \SAMPLES_i$\;
        \Comment{Train the $\NMODULES^{th}$ ART module~\cite{carpenter1991fuzzyart} (not necessary if $\LABELS \in \mathbb{N}^{N \times 1}$: $\OUTPUT_{\NMODULES} \gets \LABELS_i$)}
        $\SAMPLE_{\NMODULES} \gets f_{\NMODULES}(\SAMPLE)$\;
        $\OUTPUT_{\NMODULES} \gets$ \FTrainARTPLayer{$\text{ART}_\NMODULES$, $\SAMPLE_{\NMODULES}$, $\theta_\NMODULES^{\text{ART}}$} \;
        \Comment{Train the $(\NMODULES - 1)$ remaining ART and Map Field modules}
        \For{
            $\LAYERINDEX = \{\NMODULES-1, \ldots, 1\}$
        }{
            \Comment{Train the $\LAYERINDEX^{th}$ simplified ARTMAP~\cite{kasuba1993simplified}}
            $\SAMPLE_k \gets f_{\LAYERINDEX}(\SAMPLE)$\;
            $\OUTPUT_k \gets$ \FTrainARTMAPLayer{%
                $\text{ART}_{\LAYERINDEX}$, $\text{MF}_k$, $\SAMPLE_k $, $\OUTPUT_{k+1}$, $\Theta_{\LAYERINDEX}$
            }\;
        }
    }
    \Comment{Inference}
    \For{
        $i = \{ 1, \ldots, \NSAMPLES \}$
    }{
        $\SAMPLE_k \gets \SAMPLES_i$\;
        \For{
            $\LAYERINDEX = \{1, \ldots, \NMODULES\}$
        }{
            $\SAMPLE_k \gets f_{k}(\SAMPLE_k)$\;
            $\OUTPUT_{i, k} \gets$ \FInferenceARTLayer{$\text{ART}_k$, $\SAMPLE_k$, $\theta_k^{\text{ART}}$} \;
        }
    }
\end{algorithm}

\section{Limitations and Considerations}
Because of the sequential processing of \method layers during training, the selection of the order and quality of the nonlinear functions $\{f_k\}_{k=1}^{L}$ has a significant effect on the resulting hierarchy of category labels in \method's map fields.
With a fixed selection of $\{f_k\}_{k=1}^{L}$, experimentation and multiple simulations may be required to determine the most useful hierarchy of \method map field category labels.

\section{Software Implementation}
A Python implementation of the \method model is publicly available through GitHub at \url{https://github.com/NiklasMelton/AdaptiveResonanceLib} or PyPi via the \texttt{artlib} package \cite{Melton_AdaptiveResonanceLib_2024}.
An implementation in the Julia programming language is also publicly available on JuliaHub as \texttt{AdaptiveResonance.jl} \cite{Petrenko2022} and GitHub at \url{https://github.com/AP6YC/AdaptiveResonance.jl}.

\section{Acknowledgments}
This research was sponsored by the Army Research Laboratory and was accomplished under Cooperative Agreement Number W911NF-22-2-0209.
The views and conclusions contained in this document are those of the authors and should not be interpreted as representing the official policies, either expressed or implied, of the Army Research Laboratory or the U.S. Government.
The U.S. Government is authorized to reproduce and distribute reprints for Government purposes notwithstanding any copyright notation herein.

Support from the Missouri S\&T Mary K. Finley Endowment and the Kummer Institute for Artificial Intelligence and Autonomous Systems is gratefully acknowledged.
This project was also supported by NSF Award Number 2420248 under the project title \textit{EAGER: Solving Representation Learning and Catastrophic Forgetting with Adaptive Resonance Theory}.

\bibliographystyle{IEEEtran}

\bibliography{
    bibs/bib.bib
}


\end{document}